\documentclass[letterpaper, 10 pt, conference]{ieeeconf}  

\IEEEoverridecommandlockouts                              

\usepackage{mars}
\usepackage{algorithmic}
\usepackage{algorithm}
\usepackage{amssymb,amsmath}
\usepackage{roboticsMathNotations}
\usepackage{xcolor}
\usepackage{url}
\usepackage{multicol}
\usepackage{multirow}
\usepackage[utf8x]{inputenc}
\usepackage[english]{babel}
\usepackage{algorithmic}
\usepackage{algorithm}
\usepackage{subfigure}
\usepackage{siunitx}
\usepackage{caption}
\usepackage{tabularx}
\usepackage{booktabs}

\newcolumntype{Y}{>{\raggedleft\arraybackslash}X}
\newcolumntype{b}{Y}
\newcolumntype{z}{>{\hsize=.55\hsize}Y}
\newcolumntype{s}{>{\hsize=.15\hsize}Y}

\makeatletter
\newcommand{\todolist}[1]{\begin{itemize}\item \textcolor{red}{#1}\checknextarg}
	\newcommand{\checknextarg}{\@ifnextchar\bgroup{\gobblenextarg}{\end{itemize}}}
\newcommand{\gobblenextarg}[1]{ \item \textcolor{red}{#1}\@ifnextchar\bgroup{\gobblenextarg}{\end{itemize}}}

\usepackage{pgf}
\newcommand\inputpgf[2]{{
		\let\pgfimageWithoutPath\pgfimage
		\renewcommand{\pgfimage}[2][]{\pgfimageWithoutPath[##1]{#1/##2}}
		\input{#1/#2}
}}
\usepackage{adjustbox}

\title{\LARGE \bf 
	Pose Estimation for Omni-directional \\ Cameras using Sinusoid Fitting
}

\author{Haofei Kuang$^{1}$, Qingwen Xu$^{1}$, Xiaoling Long$^{1}$ and S\"oren Schwertfeger$^{1}$
\thanks{$^{1}$All authors are with the School of Information Science Technology of ShanghaiTech University, China
	{\tt\small <kuanghf, xuqw, longxl, soerensch>@shanghaitech.edu.cn}}%
}

\begin{document}
	
%
%


\marsPublishedIn{Accepted for:} 		

\marsVenue{IEEE/RSJ International Conference on Intelligent Robots and Systems (IROS) 2019}

\marsYear{2019}

\marsPlainAutors{Haofei Kuang, Qingwen Xu, Xiaoling Long and S\"oren Schwertfeger}


\marsMakeCitation{Pose Estimation for Omni-directional Cameras using Sinusoid Fitting}{IEEE Press}

\marsDOI{\url{}}

\marsIEEE{}


\makeMARStitle

%
%

\maketitle
\begin{abstract}
	
		We propose a novel pose estimation method for geometric vision of omni-directional cameras. On the basis of the regularity of the pixel movement after camera pose changes, we formulate and prove the sinusoidal relationship between pixels movement and camera motion. We use the improved Fourier-Mellin invariant (iFMI) algorithm to find the motion of pixels, which was shown to be more accurate and robust than the feature-based methods. While iFMI works only on pin-hole model images and estimates 4 parameters (x, y, yaw, scaling), our method works on panoramic images and estimates the full 6 DoF 3D transform, up to an unknown scale factor. For that we fit the motion of the pixels in the panoramic images, as determined by iFMI, to two sinusoidal functions. The offsets, amplitudes and phase-shifts of the two functions then represent the 3D rotation and translation of the camera between the two images. We perform experiments for 3D rotation, which show that our algorithm outperforms the feature-based methods in accuracy and robustness. We leave the more complex 3D translation experiments for future work. 
\end{abstract}

\section{Introduction}
\label{introduction}

Omni-directional cameras have been widely used in mobile robots. Visual cues from panorama images help the robot to achieve homing in \cite{argyros2005robot}, which uses omni-cameras' advantage of the large field of view (FOV). In \cite{lemaire2007slam}, panorama images are exploited to implement a bearings-only SLAM system, which can provide rich feature points. In addition, several meaningful applications with panorama cameras are mentioned in \cite{benosman2000panoramic}. In the above cases, pose estimation is one of the important topics. Optimization and geometric methods are two common ways to achieve this task. The former is usually used in feature-based methods \cite{davison2007monoslam}, while the latter is one of the basic parts in direct methods \cite{engel2014lsd}. These two approaches are usually combined together to achieve better performance. For example, Engel et. al combined both methods to realize real-time robust visual odometry in \cite{engel2018direct}; \cite{mur2015orb} uses geometric methods for the front-end of a SLAM system and the optimization method for the back-end. In addition, recently deep learning gained attention in pose estimation \cite{handa2016gvnn,costante2016exploring}. Since the deep learning methods heavily depend on training images, which may be difficult to adapt to different environments within limited training datasets, we do not take the deep learning method into account.

Optimization methods are usually used to minimize the photometric and geometry errors. In direct methods based on photometric consistency, the motion between different frames is estimated by minimizing the photometric error, as used in \cite{engel2014lsd,engel2018direct, newcombe2011dtam}. Geometry error optimization helps to boost pose estimation, such as bundle adjustment \cite{triggs1999bundle} and graph optimization \cite{kummerle2011g, IJRR2016GraphSLAM}.

Geometric methods include two main groups: pose estimation by 2D to 2D correspondences and 3D to 2D correspondences. The eight-point \cite{hartley1995defence} and five-point algorithms \cite{stewenius2005minimal} are widely used to estimate the relative pose between two frames with given 2D-2D correspondences. Firstly, the essential matrix or fundamental matrix is calculated by epipolar geometry. Then relative rotation and up-to-scale translation are estimated from the matrix. Methods like \cite{zheng2013revisiting}, that estimate relative pose by 3D-2D correspondences, compensate the scaling case and calculate the absolute pose directly, because 3D points provide the real scale information.

\begin{figure}
	\centering
	\subfigure[Low-cost exploratory robot]{\label{fig:demo robot}
		\includegraphics[height=0.35\columnwidth]{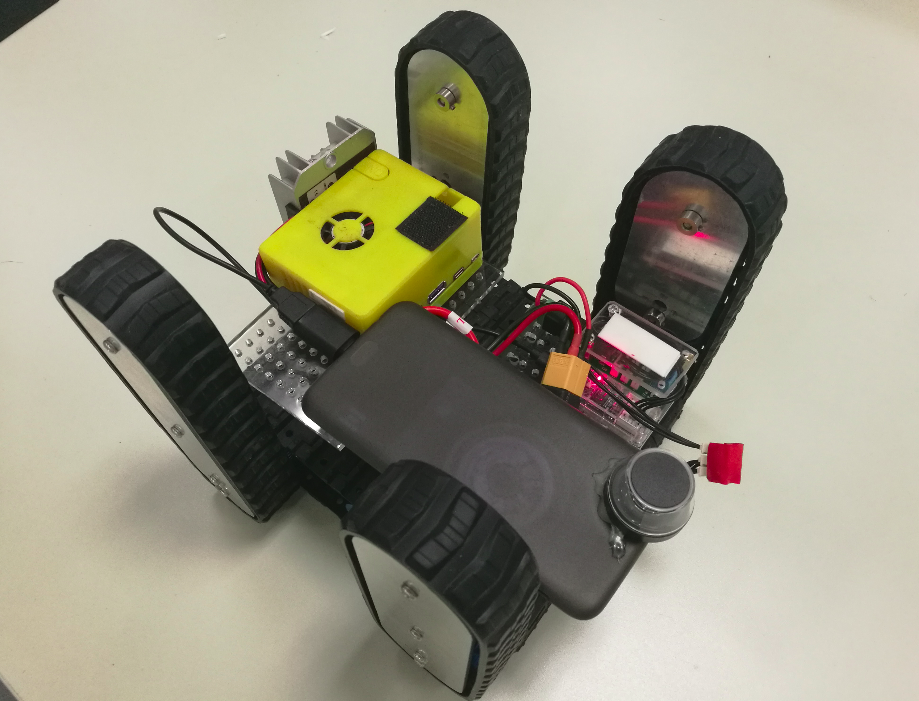}
	}
	\subfigure[Fisheye lens image]{\label{fig:fisheye lens}
		\includegraphics[height=0.35\columnwidth]{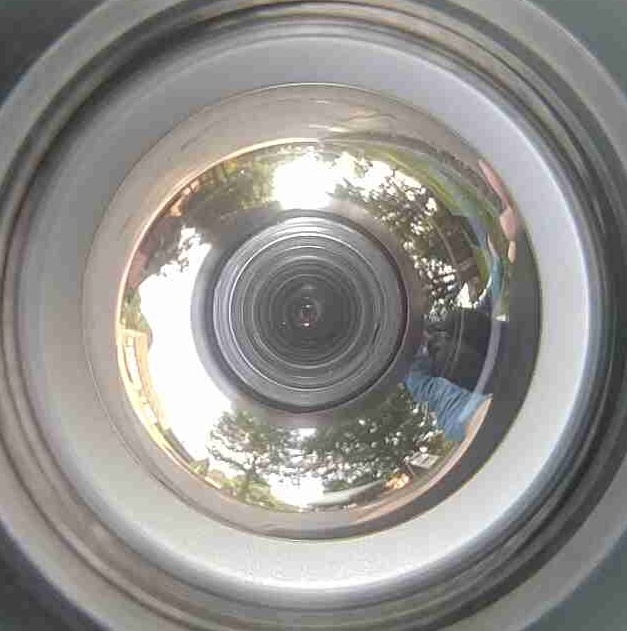}
	}
	\\
	\subfigure[Panorama image ($110$px $\times$ $1100$px)]{\label{fig:omni lens}
		\includegraphics[width=0.9\columnwidth]{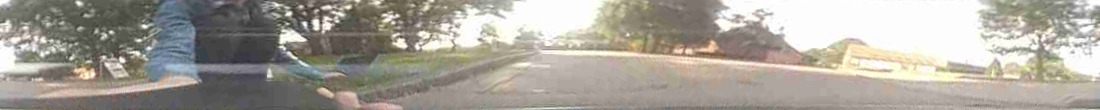}
	}
	\captionsetup{justification=justified}
	\caption{A tracked robot with a smartphone as control unit. An omni-directional lens is placed over the front-camera of the phone. The captured image as well as the unwrapped panoramic image are shown.}
	\label{fig:system setup}
\end{figure}

In addition to feature-based and direct methods, spectral methods are also used for motion estimation. On 2D images, the iFMI algorithm \cite{bulow2009fast, iFMI-Resolution-IAV10} uses Fast Fourier Transform (FFT) to calculate the $x$ and $y$ translation, yaw and scaling between two images. It is applied to image mosaicking \cite{JIRS-UAVmosaic-ELROB-RREE-2011} and motion detection \cite{SSRR11-UAV-MotionDetection}. A related approach is also used for registration of 3D range data \cite{PAMI2013Spectral}.

If we only consider pose estimation between two frames, optimization is used with direct methods, while the geometric vision helps feature-based methods to calculate the camera's motion. Our proposed method belongs to the latter, but it differs from the two approaches mentioned above by directly exploiting the properties of catadioptric omni-directional cameras. Figure~\ref{fig:system setup} shows the omni-directional camera that we use to collect omni-images and one omni-image, mounted on a simple robot. We also show the image as captured by the smart phone and its unwrapped panorama image. From Figure~\ref{fig:omni lens}, we can see that the catadioptric panorama image has the disadvantage of low resolution, as mentioned in \cite{benosman2000panoramic}, which will increase the difficulty for feature-based methods. Thus it is important to process these images with more robust algorithms. To meet this requirement, we combined the improved Fourier-Mellin invariant (iFMI) algorithm with our proposed method, since it proves to be more robust than SIFT.
Our main contributions are summarized as:
\begin{enumerate}
	\item proposing a novel relative pose estimation method based on geometric vision and fitting of pixel displacement values to sinusoidal functions;
	\item exploiting the 2D frequency-based algorithm to estimate 3D pose of the omni-camera;
	\item comparing our algorithm with commonly used epipolar geometry methods together with different features.
\end{enumerate}

The rest of this paper is organized as follows: the theoretical foundation of our proposed method is introduced in Sec~\ref{sec:problem_formulation}, including motion model, proof and parameter estimation; then we explain out implementation detail in Sec~\ref{sec:implementation}; the experiment and analysis is displayed in Sec~\ref{sec:experiments}; finally, we conclude our work in Sec~\ref{sec:conclusion}.


\section{Problem Formulation}
\label{sec:problem_formulation}

\subsection{Notations}
\label{ssec:notations}
\begin{itemize}
	\item \textit{u}-,\textit{v}-axis: image coordinates
	\item \textit{x}-,\textit{y}-,\textit{z}-axis: camera coordinates
	\item $I_{o}$: an omni-image; $^kI_{o}$: superscript $k$ means $k^{th}$frame
	\item $I_{p}$: a panorama image
	\item $u_p, v_p$: coordinates in a panorama image;
	\item $u_{\max}, v_{\max}$: width and height of the panorama image;
	\item $\Delta u(u_p)$: the translation (in pixel) in \textit{u}-axis versus a column ($u_p$) between two panorama images;
	\item $\Delta v(u_p)$: the translation (in pixel) in \textit{v}-axis versus a column ($u_p$) between two panorama images;
	\item $\mathrm{R}$: 3D rotation of the camera between two panorama images;
	\item $\mathrm{t}$: 3D translation of the camera between two panorama images;
	\item $\mathrm{P_{xy}(t)}$: projection of $t$ in \textit{x-y} plane;
	\item $\mathrm{P_{z}(t)}$: projection of $t$ in \textit{z} axis;
	\item $\mathrm{P_{xy}(R)}$: projection of $R$ around \textit{x} and \textit{y} axis;
	\item $\mathrm{P_{z}(R)}$: projection of $R$ around \textit{z} axis;
	\item $\hat{\mathrm{t}}_{xy}$: the angle between the axis of the \textit{x},\textit{y} translation and \textit{x} axis;
	\item $\lambda$: the unknown scale factor of the translation;
	\item $\gamma$: the known opening angle of a pixel. Suppose pixels are square, then $\gamma = \frac{2 \pi }{u_{\max}}$;
	\item $r$: the radius of the cylinder;
	\item $H$: the height of the cylinder; thus the height of corresponding panorama image;
	\item $T$: transformation matrix in 3D space.
\end{itemize}

\subsection{Modeling of Pose Estimation}

\subsubsection{Cylinder Camera Model}
\begin{figure}[tb]
	\centering
	\includegraphics[width=0.9\linewidth]{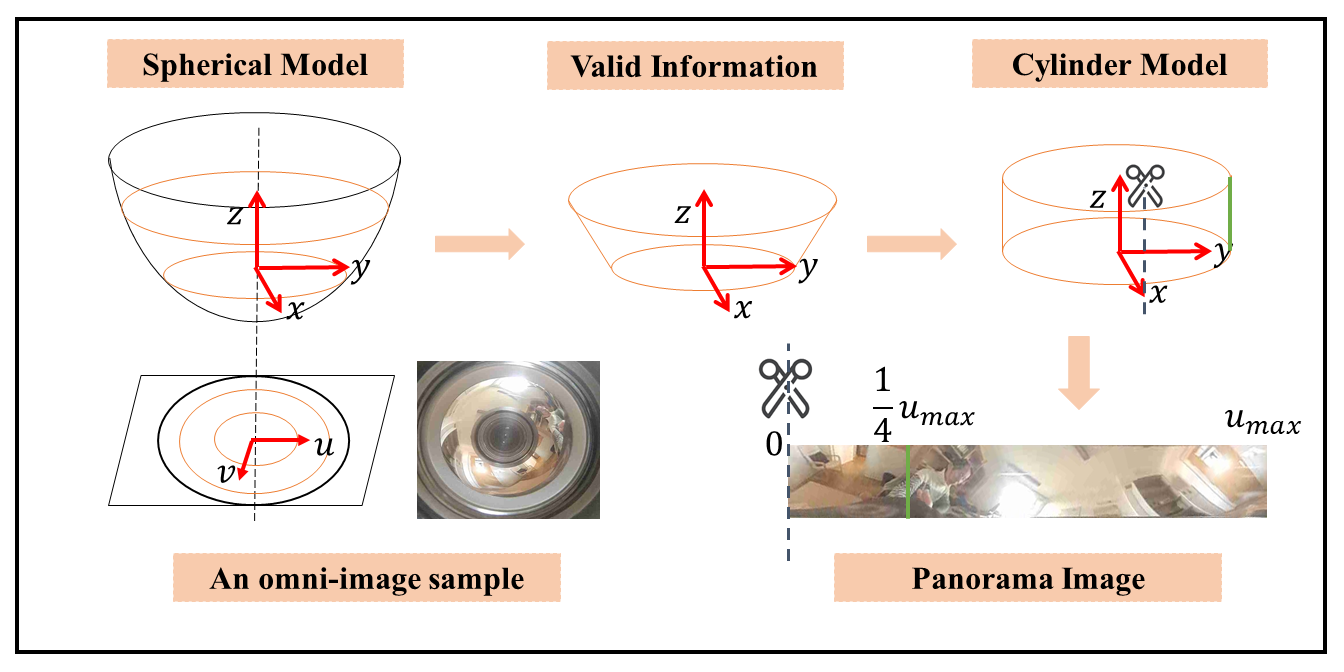}
	\caption{Cylindric Camera Model}
	\label{fig:cylinder_model}
\end{figure}

As mentioned in \cite{scaramuzza2014omnidirectional}, an image captured by a catadioptric omni-directional camera is usually unwrapped into a cylindric panorama image. Thus we model the pose estimation problem base on the cylindric camera model in this work. Figure~\ref{fig:cylinder_model} gives an intuitive description from spherical to cylinder model, which can be implemented by interpolation. The transformation between the cylinder and the panorama image can be described as
\begin{subequations}
	\begin{equation}
	u = r\cdot\theta = r\cdot\arctan(\frac{y}{x})
	\end{equation}
	\begin{equation}
	v = \frac{H}{2} - z
	\end{equation}
	\label{eq:cylinder2pano}
\end{subequations}
and
\begin{subequations}
	\begin{equation}
	x = r\cdot\cos\theta = r\cdot\cos\frac{u}{r}
	\end{equation}
	\begin{equation}
	y = r\cdot\sin\frac{u}{r}
	\end{equation}
	\begin{equation}
	z = \frac{H}{2} - v
	\end{equation}
	\label{eq:mapping_pano_cylinder}
\end{subequations}
Note that in the panorama image there will be a $u$ associated with the positive $x$-axis of the camera (e.g. front of the robot) and another $u$, $\frac{1}{4} u_{max}$ pixels to the right, which is associated with the positive $y$-axis of the camera (e.g. left of the robot). The negative sides of the axes will be on the opposite side of the panorama image, so $\frac{1}{2} u_{max}$ pixels away. In order to simplify the formulation, this paper assumes that the x-axis will always located at $u=0$ (grey dash line in Figure~\ref{fig:cylinder_model}), and the y-axis is thus at $\frac{1}{4} u_{max}$ (green line in Figure~\ref{fig:cylinder_model}).

\subsubsection{Resolution Consistency}
When unwrapping omni-images to panorama images with Eq.~\ref{eq:cylinder2pano},~\ref{eq:mapping_pano_cylinder}, we cannot make sure that each pixel is square, i.e. there may be resolution inconsistency. In other words, the incident angle could be different with same pixels in $u-$ and $v-$axis. Thus we use a calibration method described in \cite{conroy1999resolution} to find the ratio between angles per pixel in $u-$ and $v-$ direction. For the remainder of this paper we assume a perfectly calibrated cylinder model with square pixels, while the detailed discussion of such calibration is out of the scope of this paper.

\subsubsection{Motion Model}
We propose a sinusoidal function to describe the motion model of catadioptric omni-directional cameras as follows,:
\begin{subequations}
	\begin{equation}
		y = B + A\sin(\omega x + \phi)
		\label{eq:motion_model_a}
	\end{equation}
	\begin{equation}
		\Delta v(u_p) = \lambda \mathrm{t}_z  + \gamma \Vert  \mathrm{P_{xy}(R)}  \Vert \cdot \sin \left( \gamma u_p +  \frac{\mathrm{P_{xy}(R)}}{\Vert  \mathrm{P_{xy}(R)}  \Vert} \right)
		\label{eq:delta_v}
	\end{equation}
	\begin{equation}
		\Delta u(u_p) = \gamma \mathrm{P_{z}(R)}  + \lambda \Vert  \mathrm{P_{xy}(t)}  \Vert \cdot \sin \left( \gamma u_p + \hat{\mathrm{t}}_{xy} \right)
		\label{eq:delta_u}
	\end{equation}
	\label{eq:motion_model}
\end{subequations}
We model the motion of the camera using the sinusoid function Eq~\ref{eq:motion_model_a} with offset $B$, amplitude $A$ and phase shift $\phi$, while $\omega$ is just a calibration factor (field of view of a pixel). In order to recover the full six degree of freedom (DoF) motion of the camera we apply this function twice: once to the vertical motion $\Delta v(u_p)$ of the columns of the panorama image (Eq.~\ref{eq:delta_v}) and once to the horizontal motion $\Delta u(u_p)$  of the columns of the panorama image (Eq.~\ref{eq:delta_u}).

Then we analyze the motion model in four different cases to explain why we choose the sinusoidal function to represent the model:
\begin{enumerate}
	\item \textbf{Translation along \textit{z}-axis ($\mathrm{t}_z$)}:  the shift along \textit{v}-axis of each column $u_p$, which is the same for all $u_p$;

	\item \textbf{Rotation around \textit{x}- and \textit{y}- axis ($R_{xy}$)}: (Roll and pitch, respectively, \textit{using pitch=0 and some roll as example}) the closer $u_p$ is to the (positive or negative) \textit{x}-axis of the cylinder model, the smaller the shift along the \textit{v}-axis; the closer $u_p$ is to the (positive or negative) \textit{y}-axis, the larger the shift along the \textit{v}-axis;

	\item \textbf{Rotation around \textit{z}-axis ($R_z$)}: (Yaw) the shift along the \textit{u}-axis in each column $u_p$, which is the same for all $u_p$;

	\item \textbf{Translation along \textit{x}- and \textit{y}-axis ($t_{x,y}$)}: (\textit{using translation along \textit{x}-axis as an example}) the closer $u_p$ is to the (positive or negative) \textit{x}-axis of the cylinder model, the smaller the shift along \textit{u}-axis of the center in each column; the closer $u_p$ is to the (positive or negative) \textit{y}-axis, the larger the shift along the \textit{u}-axis.
\end{enumerate}

When the camera rotates around an arbitrary axis $\vec{c}$ in the \textit{x-y} plane, the shift along \textit{v} of the columns $u_p$ follows a certain pattern: the closer the column $u_p$ is to $\vec{c}$, the smaller the shift is. Thus we use the phase shift $\frac{\mathrm{P_{xy}(R)}}{\Vert  \mathrm{P_{xy}(R)}  \Vert}$ to model the angle of $\vec{c}$ to the \textit{x}-axis in Eq~\ref{eq:delta_v}. The phase shift $\hat{\mathrm{t}}_{xy}$  in Eq~\ref{eq:delta_u} is derived following the same argument.

In the following we give a rigorous proof of the motion model. We have two images  $I_{p1}$ and  $I_{p2}$ and a transfrom $(R, t)$ between them. Assume there is an arbitrary point ${^2p_P} = [u_{p2}, v_{p2}]$ in the second panorama image $I_{p2}$, its cylinder coordinate ${^2P}$ is
\begin{equation}
{^2P} = {\left[\begin{array}{c}
	x_2 \\
	y_2 \\
	z_2
	\end{array}\right]}= {\left[\begin{array}{c}
	r\cos\frac{u}{r} \\
	r\sin\frac{u}{r} \\
	\frac{H}{2} - v
	\end{array}\right]}
\end{equation}
Then we analyze the shifts of each row  or column when the camera moves like the above four cases. Firstly, the 3D point ${^2P}$ is transformed to ${^1P} = \left[x_1, y_1, z_1\right]^T$ with specified transformation matrix $T$; secondly, we  find the intersection ${^1P}' = \left[x_1', y_1', z_1'\right]^T$ between the vector $O{^1P}$ and the cylinder $\{C: x^2 + y^2 = r^2\}$, which is then unwrapped into the point ${^1p_P}$ in the panorama image $I_{p1}$; finally, we calculated the shift between ${^1p_P}$ and ${^2p_P}$ in row and column direction, respectively.
\begin{enumerate}
	\item \textbf{Translation along \textit{z}-axis}:
	\begin{subequations}
		The transformation matrix $T$ is:
		\begin{equation}
			T = {\left[\begin{array}{cccc}
				1 & 0 & 0 & 0 \\
				0 & 1 & 0 & 0 \\
			    0 & 0 & 1 & t_z
				\end{array}\right]}
		\end{equation}
		the transformed point ${^1P}$ is
		\begin{equation}
			{^1P} =  {\left[\begin{array}{c}
				r\cos\frac{u}{r} \\
				r\sin\frac{u}{r} \\
				\frac{H}{2} - v + t_z
				\end{array}\right]}
		\end{equation}
		the intersection ${^1P}'$ is
		\begin{equation}
			{^1P'} = {\left[\begin{array}{c}
				r\cos\frac{u}{r} \\
				r\sin\frac{u}{r} \\
				\frac{H}{2} - v + \lambda_z t_z
				\end{array}\right]}
		\end{equation}
		finally we get the shift $\Delta v$ in column direction:
		\begin{equation}
			\Delta v = (\frac{H}{2} - z_2) - (\frac{H}{2} - z_1') = \lambda_z t_z
		\end{equation}
	\end{subequations}

	\item \textbf{Rotation around \textit{x}- and \textit{y}- axis}: (for example \textit{x}- axis (roll))

	\begin{subequations}
		The transformation matrix $T$ is:
		\begin{equation}
			T = {\left[ \begin{array}{cccc}
				1 & 0 & 0 & 0 \\
				0 & \cos\theta_x & -\sin\theta_x & 0\\
				0 & \sin\theta_x & \cos\theta_x & 0
				\end{array}
				\right]}
		\end{equation}
		the transformed point ${^1P}$ is
		\begin{equation}
			{^1P} = {\left[\begin{array}{c}
				r\cos\frac{u}{r} \\
				r\sin\frac{u}{r}\cos\theta_x - (\frac{H}{2}-v)\sin\theta_x \\
				r\sin\frac{u}{r}\sin\theta_x + (\frac{H}{2}-v)\cos\theta_x
				\end{array}\right]}
		\end{equation}
	  the intersection ${^1P}'$ is
		\begin{equation}
			{^1P'} = {\left[\begin{array}{c}
				\frac{r}{\sqrt{1+k^2}} \\
				\frac{kr}{\sqrt{1+k^2}} \\
				\frac{r\sin\frac{u}{r}\sin\theta_x+(\frac{H}{2}-v)\cos\theta_x}{r\cos\frac{u}{r}}\frac{r}{\sqrt{1+k^2}}
				\end{array}\right]}
		\end{equation}
		where $k = \frac{r\sin\frac{u}{r}\cos\theta_x-(\frac{H}{2}-v)\sin\theta_x}{r\cos\frac{u}{r}}$; finally we get the shift $\Delta v$ in coulumn direction:


		\begin{equation}
		\begin{split}
		\small
			\Delta v = & \frac{r(r\sin\frac{u}{r}\sin\theta_x + (\frac{H}{2}-v)\cos\theta_x)}{\sqrt{(r\cos\frac{u}{r})^2+(r\sin\frac{u}{r}\cos\theta_x-(\frac{H}{2}-v)\sin\theta_x)^2}}
			\label{eq: roll_shift} \\
			&   - (\frac{H}{2} - v)
		 \end{split}
		\end{equation}

		The denominator in Eq~\ref{eq: roll_shift} can expand into $r^2 - r\sin\frac{u}{r}(\frac{H}{2}-v)\sin 2\theta_x + (\frac{H}{2}-v)^{2}\sin^{2}\theta_{x}$, which can be approximated to $r^2$ when $\theta_x$ is small. Under this condition, $\sin\theta_x \approx \theta_x $ and $\cos\theta_x \approx 1$. Thus the shift will be
		\begin{equation}
		\Delta v = (\frac{H}{2} - z_2) - (\frac{H}{2} - z_1')  = \theta_xR\sin\frac{u}{R}
		\end{equation}
	\end{subequations}

	\item \textbf{Rotation around \textit{z}-axis}:

	\begin{subequations}
		The transformation matrix $T$ is:
		\begin{equation}
			T = {\left[ \begin{array}{cccc}
				\cos\theta_z & -\sin\theta_z & 0 & 0 \\
				\sin\theta_z & \cos\theta_z & 0 & 0 \\
				0 & 0 & 1 & 0
				\end{array}
				\right]}
		\end{equation}
		the transformed point ${^1P}$ is
		\begin{equation}
			{^1P} = {\left[\begin{array}{c}
				r\cos(\frac{u}{r} + \theta_z) \\
				r\sin(\frac{u}{r} + \theta_z) \\
				\frac{H}{2} - v
				\end{array}\right]}
		\end{equation}
		the intersection ${^1P}'$ remains
		\begin{equation}
			{^1P'} ={^1P} =  {\left[\begin{array}{c}
				r\cos(\frac{u}{r} + \theta_z) \\
				r\sin(\frac{u}{r} + \theta_z) \\
				\frac{H}{2} - v
				\end{array}\right]}
		\end{equation}
		finally we get the shift $\Delta u$ in row direction:
		\begin{equation}
			\Delta u = r\arctan\frac{y_2}{x_2} - r\arctan\frac{y_1'}{x_1'} = -r\theta_z
		\end{equation}
	\end{subequations}

	\item \textbf{Translation along \textit{x}- and \textit{y}-axis}: (for example along \textit{x}-axis)

	\begin{subequations}
		The transformation matrix $T$ is:
		\begin{equation}
		T = {\left[\begin{array}{cccc}
			1 & 0 & 0 & t_x \\
			0 & 1 & 0 & 0 \\
			0 & 0 & 1 & 0
			\end{array}\right]}
		\end{equation}
		the transformed point ${^1P}$ is
		\begin{equation}
			{^1P} = {\left[\begin{array}{c}
				r\cos\frac{u}{r} + \lambda t_x \\
				r\sin\frac{u}{r} \\
				\frac{H}{2} - v
				\end{array}\right]}
		\end{equation}
		the intersection ${^1P}'$ is
		\begin{equation}
			{^1P'} = {\left[\begin{array}{c}
				\frac{r}{\sqrt{1+k^2}} \\
				\frac{kr}{\sqrt{1+k^2}} \\
				\frac{\frac{H}{2}-v}{r\cos\frac{u}{r}+\lambda t_x}\frac{r}{\sqrt{1+k^2}}
				\end{array}\right]}
		\end{equation}
		where $k = \frac{r\sin\frac{u}{r}}{r\cos\frac{u}{r}+\lambda t_x}$. ; finally we get the shift $\Delta u$ in row direction:
		\begin{equation}
			\Delta u = r\arctan\frac{y_2}{x_2} - r\arctan\frac{y_1'}{x_1'}
		\end{equation}
		\begin{equation}
			= r\arctan\frac{\lambda t_x\sin\frac{u}{r}}{\lambda t_x\cos\frac{u}{r} + r}
		\label{eq:shift_tx}
		\end{equation}
		In Eq~\ref{eq:shift_tx}, since $r >> \lambda t_x\cos\frac{u}{r}$, then $\lambda t_x\cos\frac{u}{r} + r$ approximately equals to $r$  and $\frac{\lambda t_x\sin\frac{u}{r}}{\lambda t_x\cos\frac{u}{r} + r}$ is small enough to take the first-order Taylor appromation. Thus, the shift of each row can be described as:
		\begin{equation}
		\Delta u \approx r\frac{\lambda t_x\sin\frac{u}{r}}{r} = \lambda t_x\sin\frac{u}{r}
		\end{equation}
	\end{subequations}
\end{enumerate}

Based on the proof, we can tell that Eq~\ref{eq:delta_u} and Eq~\ref{eq:delta_v} is correct when the translation is small enough  in \textit{x}-\textit{y} plane or the rotation component along axis in \textit{x}-\textit{y} plane is small enough.

\subsection{Curve Fitting}
There are some unknown parameters $\Phi = \{ A, \phi, B \}$ in Eq.~\ref{eq:motion_model} that we estimate using curve fitting. We solve it through modeling it as a nonlinear least-squared problem and the parameters are estimated by using a nonlinear optimization algorithm. Then the rotation $R$ and translation $t$ can be calculated from $\Phi$.

To find the corresponding parameters $\Phi_v, \Phi_u$ in Eq.~\ref{eq:delta_v} and Eq.~\ref{eq:delta_u} we build the following two objective functions:
\begin{subequations}
	\begin{align}
		\label{eq:cost_v}
		C_v(u_p, \Phi_v) &= \Delta v(u_p; \Phi_v) - y_v\\
		\label{eq:cost_u}
		C_u(u_p, \Phi_u) &= \Delta u(u_p; \Phi_u) - y_u\\
		\label{eq:obj_v}
		\min_{\Phi_v} L_v(u_p; \Phi_v) &= \min_{\Phi_v} \frac{1}{2}\left\Vert C_v(u_p, \Phi_v)\right\Vert^{2}_{2}\\
		\label{eq:obj_u}
		\min_{\Phi_u} L_u(u_p; \Phi_u) &= \min_{\Phi_u} \frac{1}{2}\left\Vert C_u(u_p, \Phi_u)\right\Vert^{2}_{2}
	\end{align}
	\label{eq: object_func}
\end{subequations}
where $ C_v $ and $ C_u $ are cost functions and $y_v$ and $y_u$ are the measured shifts in $v-$ and $u-$ direction by using the iFMI algorithm \cite{JIRS-UAVmosaic-ELROB-RREE-2011}. $ L_v $ and $ L_u $ are loss functions in the standard least-squared form. Afterwards, the nonlinear optimization Levenberg-Marquardt algorithm \cite{marquardt1963algorithm} is used to minimize the objective functions, which could also be replaced with other optimization methods.

\begin{subequations}
	\begin{align}
	\label{eq:robust_obj_v}
	L_v(u_p; \Phi_v, \delta) &= \delta \cdot \left(\sqrt{1 + \left(\frac{C_v(u_p, \Phi_v)}{\delta}\right)^2} - 1\right)\\
	\label{eq:robust_obj_u}
	L_u(u_p; \Phi_u, \delta) &= \delta \cdot \left(\sqrt{1 + \left(\frac{C_u(u_p, \Phi_u)}{\delta}\right)^2} - 1\right)
	\end{align}
	\label{eq: robust_object_func}
\end{subequations}

We use two methods to handle the outliers problem. First we employ a median filter of data to remove some obvious outliers. Then we use the Huber loss function to reduce the influence of outliers, which was proven to be less sensitive to outliers in data than the L2 loss~\cite{huber1992robust}. Because the Pseudo-Huber loss function~\cite{charbonnier1997deterministic} combines the best properties of L2 loss and Huber loss by being strongly convex when close to the minimum and less steep for outliers, we use it to replace the standard least-squared loss form with Eq.~\ref{eq:robust_obj_v} and Eq.~\ref{eq:robust_obj_u}.

\section{Implementation}
\label{sec:implementation}
The implementation of our proposed method is described in Algorithm~\ref{alg:ifmi}, where we have $W = 1100, L = H =110, d = 20$.

\begin{figure}[tb]
	\centering
	\subfigure[Frame 1]{
		\label{fig:impl:pitch2}
		\includegraphics[width=1\linewidth]{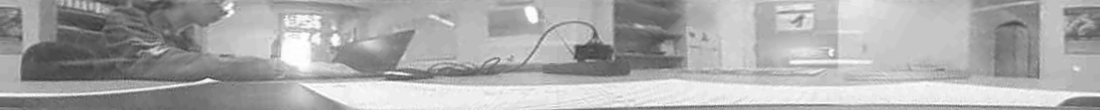}}
	\subfigure[Frame 2]{
		\label{fig:impl:pitch6}
		\includegraphics[width=1\linewidth]{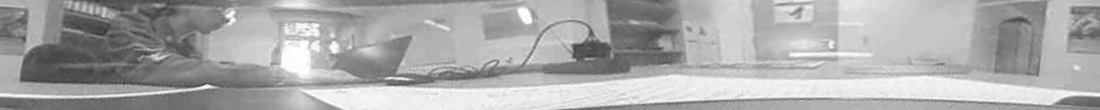}}
	\subfigure[Fitting Results]{
		\label{fig:impl:curve}
		\includegraphics[width=1\linewidth]{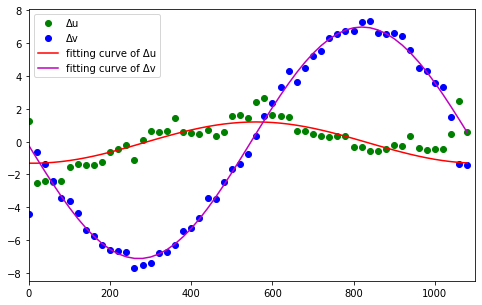}}
	\caption{$ \Delta u $ and $ \Delta v $ is shift of u and v direction, respectively, which are measured using iFMI. The results of function fitting by non-linear least squared method through the $ \Delta u $ and $ \Delta v $ values are shown. The motion corresponds to a roll (column 0 is the $x-$axis).}
	\label{fig:impl:fitting}
\end{figure}

\begin{algorithm}[b!]
	\small
	\caption{Proposed pose estimation for omni-cameras}
	\label{alg:ifmi}
	\begin{algorithmic}[1]
		\STATE \textbf{Input:} Omni images $I_{o1}$, $I_{o2}$; \\ Sliding window size $L\times L$ and step $d$
		\STATE Obtain panorama images $I_{p1}$, $I_{p2}$ of size $W \times H$\\ by cartesian-to-polar transformation
		\FOR{$L+k\times d \leq W, k\in \mathbb{N} $}
		\STATE Compute the scaling, rotation and translation $t_{uv}$ \\ for $k^{th}$ window between $I_{p1}$ and $I_{p2}$ by iFMI\cite{bulow2009fast}
		\STATE Push $t_{uv}$ to motion set $\mathbb{M}$
		\ENDFOR
		\STATE Estimate parameter $\Phi_v$ and $\Phi_u$ by optimization on $\mathbb{M}$ (Eq.~\ref{eq: object_func})
		\STATE Calculate transformation $T$ from $\Phi_v, \Phi_u$ (Eq.~\ref{eq:motion_model})
		\STATE \textbf{Output:} $T$
	\end{algorithmic}
\end{algorithm}

We use a square window to slide along the $u-$direction on the panorama images $I_{p1}$ and $I_{p2}$. For each window pair, we use the iFMI method to find the 2D motion. Then we get the set of shifts $\Delta u$ and $\Delta v$ versus column index $u_p = \frac{1}{2}L + k\times d $.
Afterwards, we fit the values with the sinusoidal function to estimate parameter $\Phi$ as line 7 of Algorithm~\ref{alg:ifmi} describes. $ \delta = 2 $ of Eq~\ref{eq: robust_object_func} is selected in our implementation. Figure~\ref{fig:impl:curve} displays an example of curve fitting between two frames.  The optimization algorithm fits the sinusoids effectively.

In Figure~\ref{fig:impl:curve} we see maxima in the $\Delta v$ curve at around $\frac{1}{4}u_{max}$ and $\frac{3}{4}u_{max}$. Following the definition from above, that the $x$-axis is at $u=0$, this means there is a big motion in the image where the $y$-axis is. It thus follows that there has been a rotation around the $x$-axis of the camera (roll).

\section{experiments}
\label{sec:experiments}
\begin{figure}[tb]
	\centering
	\subfigure[Grass datasets]{
		\label{fig:office_dataset}
		\includegraphics[height=0.3\linewidth]{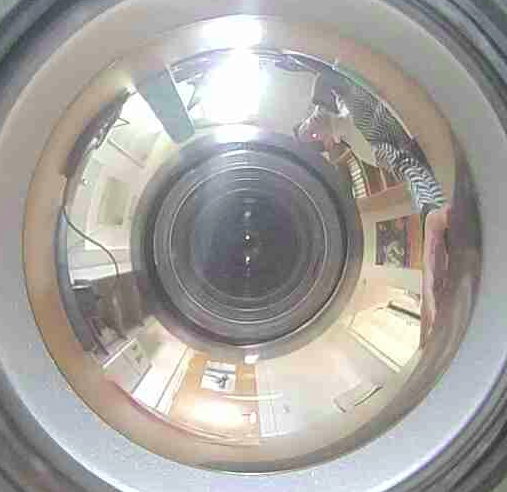}}
	\subfigure[Office datasets]{
		\label{fig:lawn_dataset}
		\includegraphics[height=0.3\linewidth]{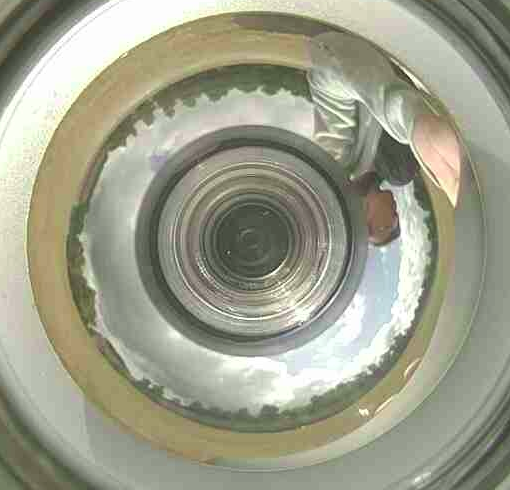}}
	\subfigure[Street datasets]{
		\label{fig:libomni_dataset}
		\includegraphics[height=0.3\linewidth]{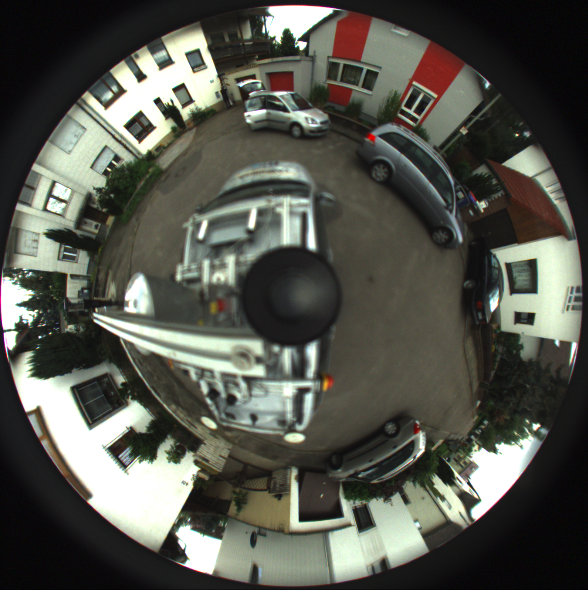}}
	\caption{Experimental Image Samples}
	\label{fig:dataset_samples}
\end{figure}

In this section we compare our method with feature-based algorithms using experiments on three different datasets. We use images from indoor (office) and outdoor (street) scenes as well as another dataset with ambiguous features (grass), which poses big challenges to feature-matching algorithms (Samples are shown in Figure~\ref{fig:dataset_samples}). 
All the images are captured by the phone (Oneplus 5) covered with a low-cost omni-lens (Kogeto Dot Lens), as shown in Figure~\ref{fig:demo robot}. In addition,
all the computations are conducted on a PC with an Intel Core i7-6700 CPU and 16 GB memory.

\begin{table*}[tb]
	\vspace{0.5cm}
	\centering
	\captionsetup{justification=centering}
	\begin{tabular}{cccccccccccccccccc}
		\toprule
		\multirow{2}{*}{} & \multicolumn{3}{c}{grass} & & \multicolumn{3}{c}{office} & & \multicolumn{3}{c}{street} & & \multicolumn{4}{c}{average error} &  \\
		& roll & pitch & yaw & & roll & pitch & yaw & & roll & pitch & yaw & & roll & pitch & yaw & $ \epsilon[rad]$ &  time[s]\\
		\cmidrule{2-4} \cmidrule{6-8} \cmidrule{10-12} \cmidrule{14-17}
		ORB & 0.196 & 0.683 & 0.239 & & 0.225 & 0.163 & 0.191 & & 0.313 & 0.339 & 0.328 & & 0.245 & 0.395 & 0.253 & 0.300 &  \textbf{0.11} \\
		\midrule
		AKAZE & 0.130 & 0.414 & \textbf{0.113} & & 0.171 & 0.110 & 0.073 & & 0.164 & 0.312 & \textbf{0.106} & & 0.155 & 0.279 & \textbf{0.097} & 0.177 &  0.75\\
		\midrule
		Ours & \textbf{0.088} & \textbf{0.123} & 0.129 & & \textbf{0.054} & \textbf{0.047} & \textbf{0.021} & & \textbf{0.027} & \textbf{0.032} & 0.143 & & \textbf{0.056} & \textbf{0.067} & 0.098 & \textbf{0.074} &  0.40 \\ \bottomrule
	\end{tabular}
	\caption{RMSE and average run-time of our method and feature-based approaches.}
	\label{tab:exp}
\end{table*}

\begin{figure*}[tb]
	\centering
	\subfigure[Grass datasets]{
		\label{fig:exp:grass}
		\includegraphics[width=0.32\linewidth]{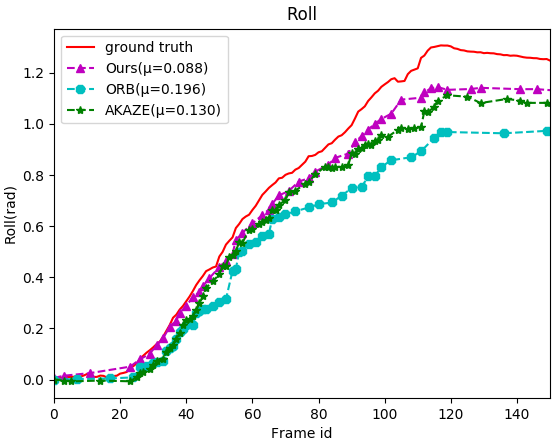}
		\includegraphics[width=0.32\linewidth]{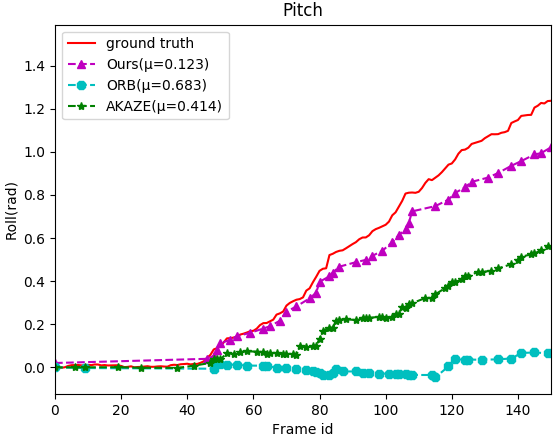}
		\includegraphics[width=0.325\linewidth]{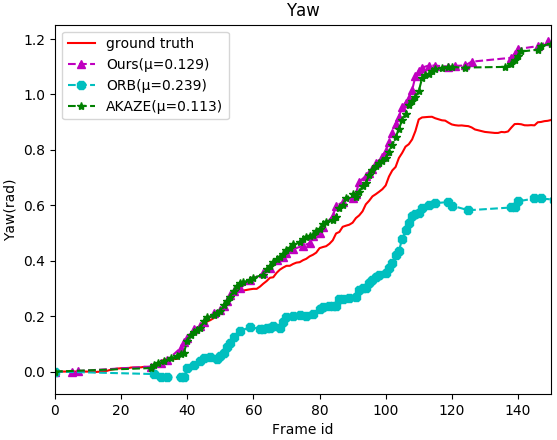}}\\
	\subfigure[Office datasets]{
		\label{fig:exp:office}
		\includegraphics[width=0.32\linewidth]{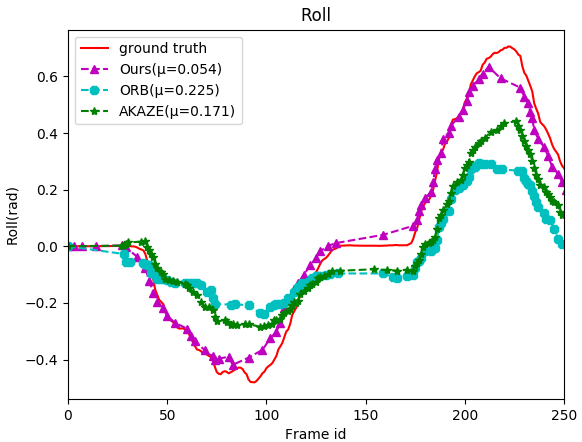}
		\includegraphics[width=0.32\linewidth]{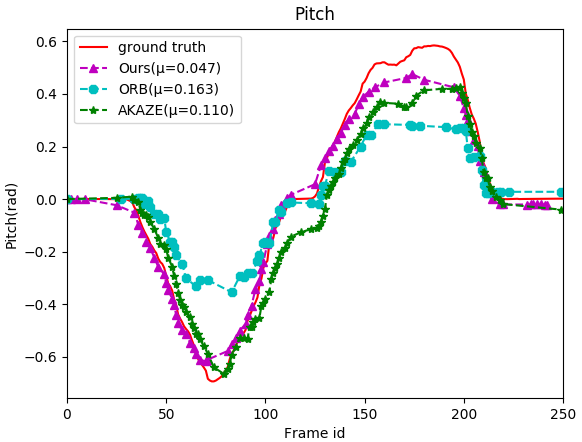}
		\includegraphics[width=0.32\linewidth]{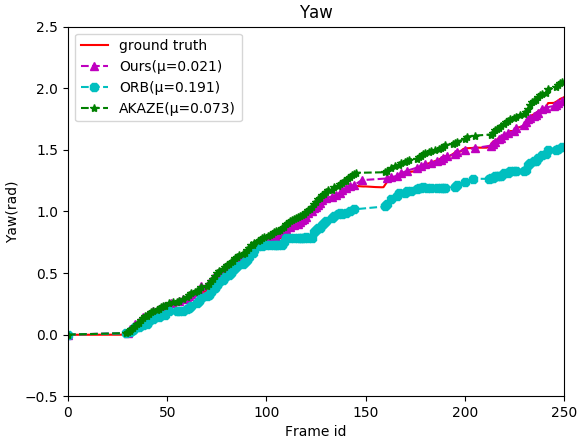}}\\
	\subfigure[Street datasets]{
		\label{fig:exp:street}
		\includegraphics[width=0.32\linewidth]{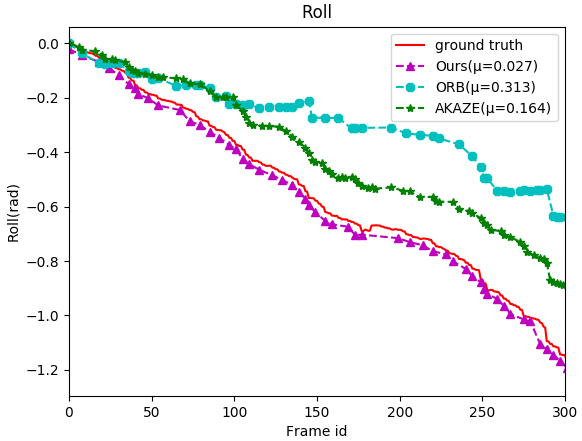}
		\includegraphics[width=0.32\linewidth]{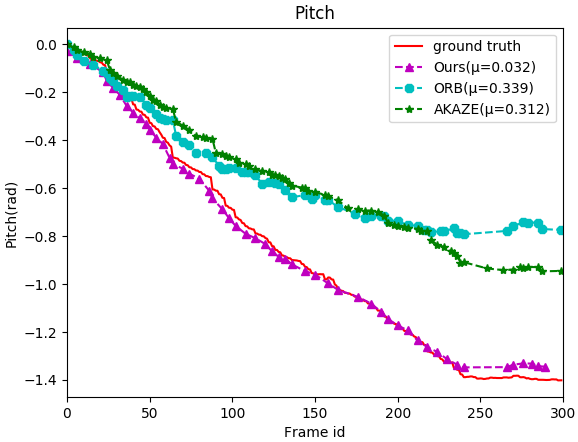}
		\includegraphics[width=0.328\linewidth]{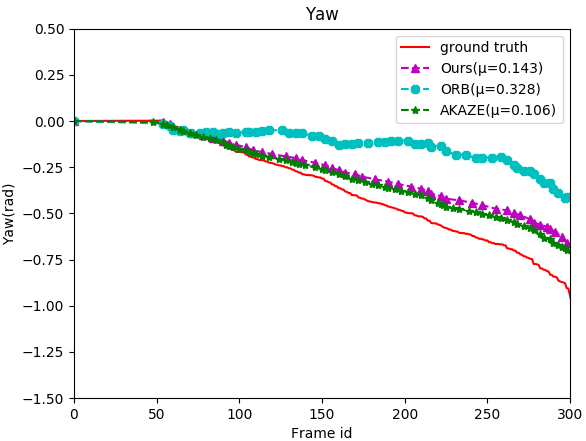}}\\
	\caption{The rotation estimation experiments with different datasets. The $\mu$ in the legend represents RMSE for each method.}
	\label{fig:exp}
\end{figure*}

In this experiment, we choose ORB and AKAZE feature detectors to implement the feature-based algorithm, since ORB is one of the fastest detectors and AKAZE is designed for detecting features in non-linear space according to \cite{tareen2018comparative}. Besides, the STEWENIUS five-point algorithm is exploited to estimate the relative pose, which is widely used and performs robustly.

To simplify the comparison, the experiment is done with controlled variables, that is, we only rotate the camera around one axis ($x-$, $y-$ or $z-$axis) at one single test. 
Experiments for the translation are omitted in this work for two reasons: For estimating the correct translation, the sinusoid fitting model assumes an equal distance to the camera of all points. This restriction is easily broken and thus the translation estimation is error prone. The rotation doesn't require this assumption. Secondly, the translation estimate is anyways up to an unknown scale factor, which adds another difficulty. We use the IMU of the smartphone as ground truth. In the following, we analyze the results in both qualitative and quantitative ways.

\subsection{Results Analysis}
Figure~\ref{fig:exp} demonstrates an intuitive performance of the pose estimation in the three datasets. We find that our method achieves a better performance than feature-based methods in general. Thanks to the robust iFMI algorithm, our method also works well on the grass dataset, while the feature-based methods suffer from describing the ambiguous features. Especially, the ORB-based algorithm fails to detect roll on the grass datasets. Both the proposed algorithm and feature-based methods show a good performance on office datasets, where the latter gets help from rich features in this scenario, e.g. chairs, books and so on. Moreover, the proposed method also outperforms the two feature-based ones on the office dataset. Lastly, the street dataset shows that our method also performs well regarding the accuracy and robustness in the general outdoor scenario.

Table~\ref{tab:exp} provides a rigorous comparison among three methods, where the root mean square error (RMSE) is used as the error metrics. This table shows that the total average error $\epsilon$ of our proposed method is less than half of the AKAZE-based algorithm and about one-quarter of the ORB-based algorithm. It also shows that the AKAZE-based method outperforms our method with a slight advantage in the yaw test. But we suspect, that the ground truth yaw (from the smartphone IMU) might not be reliable even for the short interval between the frames and thus the yaw-experiments shouldn't be trusted too much. The run-time of the AKAZE-based algorithm is more than that of our proposed and ORB-based methods, as the last column of Table~\ref{tab:exp} shows.

In summary, the our method is the most robust among the three algorithms in the different application scenarios and performing especially well in environments which are feature-deprived or have repetitive features, such as grass. Even though our method is not the fastest one, it is cost-effective when considering the balance between accuracy and run-time.

\section{Conclusion}
\label{sec:conclusion}

In this paper we proposed a novel pose estimation method by fitting sinusoid functions of pixel displacements in panoramic images of omni-directional cameras. We calculated the average pixel displacements of columns in the image using the iFMI algorithm. Experimental results for the 3D rotation part show that our method outperforms feature-based approaches, which we attribute to the facts that we use a spectral method and then fit many results to (sinusoid) functions, which thus takes care of outliers of the spectral method. The iFMI method can only estimate 2D transforms of pin-hole images, while our approach can, in principal, estimate the full 6 DoF 3D camera pose (up to scale factor) from panoramic images.


From the experiment results, we can see that the accuracy of our proposed method is about two to four times better than the other feature-based approaches for roll and pitch. The speed of our method is almost twice as fast than the accurate AKAZE-based algorithm. Thus our method is the most cost-effective one among the three methods.

We envision our algorithm to be used more for fast visual-odometry rather than visual SLAM with big displacements between frames. In this case we estimate that also the translation part of our sinusoid fitting model will perform reasonably well. Soon we will do experiments also regarding the translation part of our algorithm, and use a tracking system to generate more precise ground truth data. Furthermore, we will attempt to incorporate also the scaling and rotation of the sliding window in our model, so we can use all results that iFMI is giving us, to maybe get even better and more robust results. We then also plan to investigate other means of generating the pixel displacements per column (e.g. optical flow, 1D spectral methods). Finally, we aim to integrate our method into a full omni-visual odometry and SLAM framework.


\bibliographystyle{IEEEtran}
\bibliography{bibliography.bib}

\begin{thebibliography}{10}
\providecommand{\url}[1]{#1}
\csname url@samestyle\endcsname
\providecommand{\newblock}{\relax}
\providecommand{\bibinfo}[2]{#2}
\providecommand{\BIBentrySTDinterwordspacing}{\spaceskip=0pt\relax}
\providecommand{\BIBentryALTinterwordstretchfactor}{4}
\providecommand{\BIBentryALTinterwordspacing}{\spaceskip=\fontdimen2\font plus
\BIBentryALTinterwordstretchfactor\fontdimen3\font minus
  \fontdimen4\font\relax}
\providecommand{\BIBforeignlanguage}[2]{{%
\expandafter\ifx\csname l@#1\endcsname\relax
\typeout{** WARNING: IEEEtran.bst: No hyphenation pattern has been}%
\typeout{** loaded for the language `#1'. Using the pattern for}%
\typeout{** the default language instead.}%
\else
\language=\csname l@#1\endcsname
\fi
#2}}
\providecommand{\BIBdecl}{\relax}
\BIBdecl

\bibitem{argyros2005robot}
A.~A. Argyros, K.~E. Bekris, S.~C. Orphanoudakis, and L.~E. Kavraki, ``Robot
  homing by exploiting panoramic vision,'' \emph{Autonomous Robots}, vol.~19,
  no.~1, pp. 7--25, 2005.

\bibitem{lemaire2007slam}
T.~Lemaire and S.~Lacroix, ``Slam with panoramic vision,'' \emph{Journal of
  Field Robotics}, vol.~24, no. 1-2, pp. 91--111, 2007.

\bibitem{benosman2000panoramic}
R.~Benosman, S.~Kang, and O.~Faugeras, \emph{Panoramic vision}.\hskip 1em plus
  0.5em minus 0.4em\relax Springer-Verlag New York, Berlin, Heidelberg, 2000.

\bibitem{davison2007monoslam}
A.~J. Davison, I.~D. Reid, N.~D. Molton, and O.~Stasse, ``Monoslam: Real-time
  single camera slam,'' \emph{IEEE Transactions on Pattern Analysis \& Machine
  Intelligence}, no.~6, pp. 1052--1067, 2007.

\bibitem{engel2014lsd}
J.~Engel, T.~Sch{\"o}ps, and D.~Cremers, ``Lsd-slam: Large-scale direct
  monocular slam,'' in \emph{European conference on computer vision}.\hskip 1em
  plus 0.5em minus 0.4em\relax Springer, 2014, pp. 834--849.

\bibitem{engel2018direct}
J.~Engel, V.~Koltun, and D.~Cremers, ``Direct sparse odometry,'' \emph{IEEE
  transactions on pattern analysis and machine intelligence}, vol.~40, no.~3,
  pp. 611--625, 2018.

\bibitem{mur2015orb}
R.~Mur-Artal, J.~M.~M. Montiel, and J.~D. Tardos, ``Orb-slam: a versatile and
  accurate monocular slam system,'' \emph{IEEE transactions on robotics},
  vol.~31, no.~5, pp. 1147--1163, 2015.

\bibitem{handa2016gvnn}
A.~Handa, M.~Bloesch, V.~P{\u{a}}tr{\u{a}}ucean, S.~Stent, J.~McCormac, and
  A.~Davison, ``gvnn: Neural network library for geometric computer vision,''
  in \emph{European Conference on Computer Vision}.\hskip 1em plus 0.5em minus
  0.4em\relax Springer, 2016, pp. 67--82.

\bibitem{costante2016exploring}
G.~Costante, M.~Mancini, P.~Valigi, and T.~A. Ciarfuglia, ``Exploring
  representation learning with cnns for frame-to-frame ego-motion estimation,''
  \emph{IEEE robotics and automation letters}, vol.~1, no.~1, pp. 18--25, 2016.

\bibitem{newcombe2011dtam}
R.~A. Newcombe, S.~J. Lovegrove, and A.~J. Davison, ``Dtam: Dense tracking and
  mapping in real-time,'' in \emph{2011 international conference on computer
  vision}.\hskip 1em plus 0.5em minus 0.4em\relax IEEE, 2011, pp. 2320--2327.

\bibitem{triggs1999bundle}
B.~Triggs, P.~F. McLauchlan, R.~I. Hartley, and A.~W. Fitzgibbon, ``Bundle
  adjustment—a modern synthesis,'' in \emph{International workshop on vision
  algorithms}.\hskip 1em plus 0.5em minus 0.4em\relax Springer, 1999, pp.
  298--372.

\bibitem{kummerle2011g}
R.~K{\"u}mmerle, G.~Grisetti, H.~Strasdat, K.~Konolige, and W.~Burgard, ``g 2
  o: A general framework for graph optimization,'' in \emph{2011 IEEE
  International Conference on Robotics and Automation}.\hskip 1em plus 0.5em
  minus 0.4em\relax IEEE, 2011, pp. 3607--3613.

\bibitem{IJRR2016GraphSLAM}
M.~Pfingsthorn and A.~Birk, ``Generalized graph slam: Solving local and global
  ambiguities through multimodal and hyperedge constraints,''
  \emph{International Journal of Robotics Research (IJRR)}, vol.~35, 2016.

\bibitem{hartley1995defence}
R.~I. Hartley, ``In defence of the 8-point algorithm,'' in \emph{Proceedings of
  IEEE international conference on computer vision}.\hskip 1em plus 0.5em minus
  0.4em\relax IEEE, 1995, pp. 1064--1070.

\bibitem{stewenius2005minimal}
H.~Stew{\'e}nius, D.~Nist{\'e}r, F.~Kahl, and F.~Schaffalitzky, ``A minimal
  solution for relative pose with unknown focal length,'' in \emph{2005 IEEE
  Computer Society Conference on Computer Vision and Pattern Recognition
  (CVPR'05)}, vol.~2.\hskip 1em plus 0.5em minus 0.4em\relax IEEE, 2005, pp.
  789--794.

\bibitem{zheng2013revisiting}
Y.~Zheng, Y.~Kuang, S.~Sugimoto, K.~Astrom, and M.~Okutomi, ``Revisiting the
  pnp problem: A fast, general and optimal solution,'' in \emph{Proceedings of
  the IEEE International Conference on Computer Vision}, 2013, pp. 2344--2351.

\bibitem{bulow2009fast}
H.~B{\"u}low and A.~Birk, ``Fast and robust photomapping with an unmanned
  aerial vehicle (uav),'' in \emph{Intelligent Robots and Systems, 2009. IROS
  2009. IEEE/RSJ International Conference on}.\hskip 1em plus 0.5em minus
  0.4em\relax IEEE, 2009, pp. 3368--3373.

\bibitem{iFMI-Resolution-IAV10}
S.~Schwertfeger, H.~B{\"u}low, and A.~Birk, ``On the effects of sampling
  resolution in improved fourier mellin based registration for underwater
  mapping,'' in \emph{7th Symposium on Intelligent Autonomous Vehicles (IAV),
  IFAC}, IFAC.\hskip 1em plus 0.5em minus 0.4em\relax IFAC, 2010.

\bibitem{JIRS-UAVmosaic-ELROB-RREE-2011}
A.~Birk, B.~Wiggerich, H.~B{\"u}low, M.~Pfingsthorn, and S.~Schwertfeger,
  ``Safety, security, and rescue missions with an unmanned aerial vehicle
  (uav): Aerial mosaicking and autonomous flight at the 2009 european land
  robots trials (elrob) and the 2010 response robot evaluation exercises
  (rree),'' \emph{Journal of Intelligent and Robotic Systems}, vol.~64, no.~1,
  pp. 57--76, 2011.

\bibitem{SSRR11-UAV-MotionDetection}
S.~Schwertfeger, A.~Birk, and H.~B{\"u}low, ``Using ifmi spectral registration
  for video stabilization and motion detection by an unmanned aerial vehicle
  (uav),'' in \emph{IEEE International Symposium on Safety, Security, and
  Rescue Robotics (SSRR)}, IEEE Press.\hskip 1em plus 0.5em minus 0.4em\relax
  IEEE Press, 2011, pp. 1--6.

\bibitem{PAMI2013Spectral}
H.~B{\"u}low and A.~Birk, ``Spectral 6-dof registration of noisy 3d range data
  with partial overlap,'' \emph{IEEE Transactions on Pattern Analysis and
  Machine Intelligence (PAMI)}, vol.~35, pp. 954--969, 2013.

\bibitem{scaramuzza2014omnidirectional}
D.~Scaramuzza, ``Omnidirectional camera,'' \emph{Computer Vision: A Reference
  Guide}, pp. 552--560, 2014.

\bibitem{conroy1999resolution}
T.~L. Conroy and J.~B. Moore, ``Resolution invariant surfaces for panoramic
  vision systems,'' in \emph{Proceedings of the Seventh IEEE International
  Conference on Computer Vision}, vol.~1.\hskip 1em plus 0.5em minus
  0.4em\relax IEEE, 1999, pp. 392--397.

\bibitem{marquardt1963algorithm}
D.~W. Marquardt, ``An algorithm for least-squares estimation of nonlinear
  parameters,'' \emph{Journal of the society for Industrial and Applied
  Mathematics}, vol.~11, no.~2, pp. 431--441, 1963.

\bibitem{huber1992robust}
P.~J. Huber, ``Robust estimation of a location parameter,'' in
  \emph{Breakthroughs in statistics}.\hskip 1em plus 0.5em minus 0.4em\relax
  Springer, 1992, pp. 492--518.

\bibitem{charbonnier1997deterministic}
P.~Charbonnier, L.~Blanc-F{\'e}raud, G.~Aubert, and M.~Barlaud, ``Deterministic
  edge-preserving regularization in computed imaging,'' \emph{IEEE Transactions
  on image processing}, vol.~6, no.~2, pp. 298--311, 1997.

\bibitem{tareen2018comparative}
S.~A.~K. Tareen and Z.~Saleem, ``A comparative analysis of sift, surf, kaze,
  akaze, orb, and brisk,'' in \emph{2018 International Conference on Computing,
  Mathematics and Engineering Technologies (iCoMET)}.\hskip 1em plus 0.5em
  minus 0.4em\relax IEEE, 2018, pp. 1--10.

\end{thebibliography}

\end{document}